\documentclass{llncs}
\usepackage{makeidx}  
\usepackage{subfig}
\usepackage{graphicx}
\usepackage{calc}
\usepackage{amssymb,amsmath,array}
\usepackage{color}

\def\argmax{\mathop{\mathrm{argmax}}}    
\def\argmin{\mathop{\mathrm{argmin}}}

\def\matX{\mathbf{X}}
\def\matY{\mathbf{Y}}
\def\matC{\mathbf{C}}
\def\matH{\mathbf{H}}
\def\vecX{\mathbf{x}}
\def\vecY{\mathbf{y}}
\def\vecC{\mathbf{c}}
\def\vecH{\mathbf{h}}

\begin{document}
\frontmatter          
\mainmatter              
\title{Decoding finger movements from ECoG signals using switching
  linear models}

%
%
\author{R\'emi Flamary and Alain Rakotomamonjy}
\authorrunning{R\'emi Flamary et al.}   
%
\tocauthor{R\'emi Flamary, Alain Rakotomamonjy}
\institute{Universit\'e de Rouen, LITIS EA 4108, Avenue de l\textquoteright Universit\'e, \\
76801 Saint-\'Etienne-du-Rouvray, France,\\
\email{\{remi.flamary,alain.rakoto\}@insa-rouen.fr}}

\maketitle              

\begin{abstract}
  One of the major challenges of ECoG-based Brain-Machine Interfaces
  is the movement prediction of a human subject. Several methods
  exist to predict an arm 2-D trajectory. The fourth BCI Competition
  gives a dataset in which the aim is to predict individual finger
  movements (5-D trajectory). The difficulty lies in the fact that
  there is no simple relation between ECoG signals and finger
  movement. We propose in this paper to decode finger flexions using 
  switching models. This method permits to simplify the system as it is
  now described as an ensemble of linear models depending on an internal
  state. We show that an interesting accuracy prediction can be obtained 
by such a model.
\end{abstract}

\section{Introduction}

Some people who suffer some neurological diseases can be highly
paralyzed because they do not have anymore control on
their muscles.  Therefore, their only way to communicate is by using
their electroencephalogram signals. Brain-Computer interfaces (BCI)
research aim at developing systems that help those disabled people
communicating with machines. Non-invasive BCIs have recently received
a lot of interest because of their easy protocol for sensors
implantation on the scalp surface \cite{Wolpaw2004,bcicometitioniii_long}.
Furthermore, although the electroencephalogram signals have been
recorded through the skull, those BCI have shown great performance
capabilities, and can be used by real Amyotrophic Lateral
  Sclerosis (ALS) patients
\cite{sellers06:_p300_based_brain_comput_inter,nijboer08:_brain_comput_inter_for_peopl}.

However, non-invasive recordings still show some drawbacks including poor
signal to noise ratio and poor spatial resolution. Hence, in order  to
overcome these difficulties, invasive BCI may be used.  For instance,
Electrocorticographic recordings (ECoG) have recently received a great
amount of interest owing to their semi-invasive nature as they are
recorded from the cortical surface. Indeed, they offer
higher spatial resolution and they are far less sensitive to artifact
noise. Feasibility of invasive-based BCI have been proven by several
recent papers
\cite{e.04:_brain_comput_inter_using_elect,hill06:_class_eeg_and_ecog_signal,hill07:_towar_brain_comput_inter,shenoy08:_gener_featur_for_elect_bci}.
In many of these papers, the BCI paradigm considered
is  motor imagery yielding thus to a binary decision BCI.

A recent breakthrough has been made by  Schalk et al. \cite{Schalk2007}
which has proven that ECoG recordings can lead
to multiple-degree BCI control. Followed by Pistohl et al.
\cite{Pistohl2008}, these two works have
considered the problem of predicting arm movements from ECoG
signals. Both approaches are based on estimating
a linear relation between features extracted
from ECoG signals and the
actual arm movement. 

In this work, we investigate a finer degree of resolution in BCI
control by addressing the problem of estimating finger flexions
through ECoG signals.  Indeed, we propose in this paper a method for
decoding finger movements from ECoG data based on switching
models. The underlying idea of switching models is the
hypothesis that movements of each of the five fingers are triggered by
an internal discrete state that can be estimated and that all finger
movements depend on that internal state.  While such an idea of
switching models have already been successfully used for arm movement
prediction on monkeys from micro-electrode array measures
\cite{darmajian06}, here, we develop a specific approach adapted to finger
movements.  The global method has been tested on the 4th Dataset of
the BCI Competition IV\cite{bcicompiv}.

The paper is organized as follows : First, we present the dataset from
the BCI Competition IV used in this paper, then we explain our
decoding method used to obtain finger flexion from ECoG signals.
Finally we present the results obtained with our method and we discuss
several ways of improving them.







\section{Dataset}
\label{sec:dataset}

For this work, the fourth dataset from the BCI Competition
IV \cite{bcicompiv} was used. The subjects were 3 epileptic patients
who had platinium electrode grids placed on the surface of their
brain. The number of electrodes vary between 48 to 64 depending on the
subject and their position on the cortex was unknown.

Electrocorticographic (ECoG)\index{ECoG} signals of the subject were
recorded at a 1KHz sampling using BCI2000 \cite{BCI2000}. A band-pass
filter from 0.15 to 200Hz was applied to the ECoG signals. The finger
flexion \index{Finger Flexion} of the subject was recorded at 25Hz and
up-sampled to 1KHz.  Due to the acquisition process, a delay appears
between the finger movement and the measured ECoG
signal.  To correct this time-lag we apply the 37 ms
delay proposed in the dataset description~\cite{bcicompiv} to the ECoG signals.

The BCI Competition dataset consists in a 10 minutes recording per
subject. 6 minutes 40 seconds (400,000 samples)  were given for the
learning models and the remaining 3 minutes 20 seconds (200,000 samples) were
used for testing. However, since the finger
flexion signals have been up-sampled and thus are partly composed
of artificial samples, we have down-sampled  the number
of points by a factor of 4 leading to a training set of size 100,000
and a testing set of size 50,000. 
The 100,000 samples provided for learning have
been splitted in a training (75,000) and validation set
(25,000). 
Then, all parameters of the approach have
been optimized in order to maximize the performance on the validation
set. 
Note that all results presented in the paper
have been obtained using the testing set  provided by the competition
(after up-sampling then back by a factor of $4$).

In this competition,  method performance was measured  through the
cross-correlation between the measured and the estimated finger
flexion. The correlation were averaged across fingers and across
subject to obtain the overall  method performance. Note that the fourth
finger was not used for evaluation in the competition since its
movements were proven to
be correlated with the other one movements~\cite{bcicompiv}.


\section{Finger flexion decoding using switching linear models}



This section presents the full methodology we have used for
addressing the problem of estimating finger flexions from
ECoG signals. In the first part, we propose an overview
of the switching models. Then, we describe 
how we learn the function that estimates which finger is
about to move. Afterwards, we detail the linear  models associated
to each moving finger. Finally, we briefly detail how the complete method
works in the decoding stage.


\subsection{Overview}
\begin{figure}[t]
 \centering
 \includegraphics[width=\linewidth]{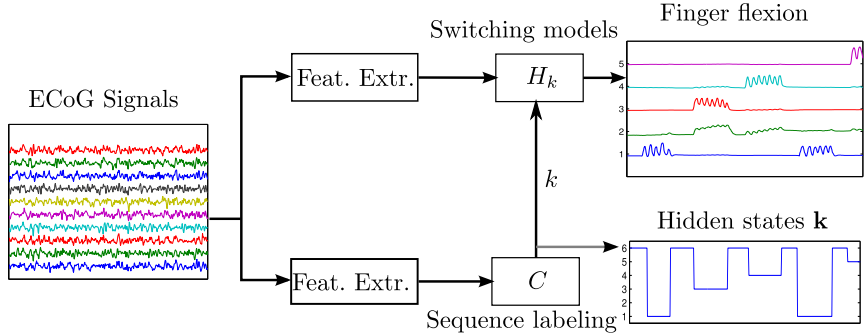}
 \caption{Diagram of our switching models decoder. We see
that from the ECoG signals, we estimate two models. (bottom flow) one
which outputs a state $k$ predicting which finger
is moving and (top flow) another one that, given the predicted
moving finger, estimates the flexion of all fingers.  }
 \label{fig:decoder}
\end{figure}

In order to obtain an efficient prediction of finger flexions, we
have made the hypothesis that for such movements the brain can be
understood as a switching model. This translates into the assumption
that the measured ECoG signals and the finger movements are
intrinsically related by an internal state $k$. In our case, this
state corresponds to each finger moving, $k=1$ for the thumb to $k=5$
for the baby finger or $k=6$ for no finger movement. Here, we used mutually-exclusive states because  the 
experimental set-up considered 
specifies that only one or no finger is moving.  
Figure \ref{fig:decoder} gives the picture of our finger movement
decoding scheme.
Basically, the idea is that based on
some features extracted from the  ECoG signals,
the internal hidden state triggering the switching finger models
can be estimated. Then, this state allows the system
to select an appropriate model $\matH_k(\tilde \vecX)$ for
estimating all finger flexions, with
$\tilde \vecX$ being a feature vector.

For the complete model, we need to estimate the function
 $f(\cdot)$ that maps the ECoG features to an internal state $k \in \{1,\cdots,6\}$
and the functions $\matH_k(\cdot)$ that relates the brain signals
to all finger flexion amplitudes.
The next paragraphs  present how we
have modeled these functions and how we have estimated
them from the data. 


\subsection{Moving finger estimation}
\label{sec:moving-fing-indent}

The methodology used for learning the $f(\cdot)$ function which estimates
the moving finger is given in the sequel. 

\subsubsection{Feature extraction}

\begin{figure}[t]
 \centering
 \includegraphics[width=\linewidth]{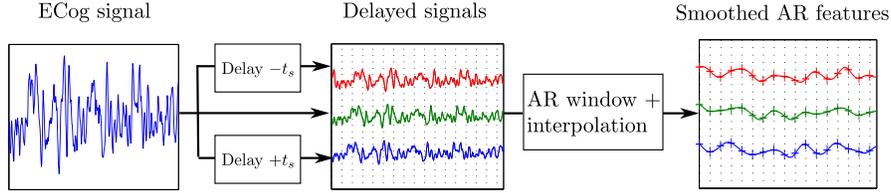}
 \caption{Diagram of the feature extraction 
procedure for the moving finger decoding. Here, we have outlined the
processing of a single channel signal.}
 \label{fig:featextr}
\end{figure}

For this problem of estimating the moving finger, the features 
we used are based on smoothed Auto-Regressive (AR) coefficient of the signal. 
The global overview of the feature extraction procedure
is given in Figure \ref{fig:featextr}. 
For a single channel, the procedure is the following. 
The signal from that channel is divided in non-overlapping
window of $300$ samples. For each window, an auto-regressive
model has been estimated. Thus, AR
coefficients are obtained at every 300 samples (denoted
by the vertical dashed line and the cross in Figure \ref{fig:featextr}).
In order to have a continuous AR coefficients value,
a smoothing spline-based  interpolation between two consecutive
AR coefficients has been used. Note that instead of interpolating,
we could have computed the AR coefficients at each time instant,
however, the approach we propose here has the double advantage
of being less-computationally demanding and  of providing
some smoothed (and thus more robust to noise) AR coefficients.
Finally, only the two first AR coefficients are used as 
features. Signal dynamics have been taken into account
by applying a similar procedure to shifted 
version of the signal at ($+t_s$ and $-t_s$).
 Hence, for measurements
involving $48$ channels, the feature vector at 
a time instant $t$ is obtained
by concatenating features extracted from
all channels, leading to a resulting vector of size $48 \times 3 \times 2 =240$.

\subsubsection{Channel Selection}
Actually, we do not consider in the model
all the channels. Indeed, a  channel selection algorithm 
has been run in order
to reduce the number of channels.
For this channel selection procedure,
the feature vector $\vecX_t$ at time $t$ has been computed
as described above, except that we have not considered
the shifted signal versions and used only
the first AR coefficient.

Then, for each finger, based on the training
set, we estimated a linear regression 
$y = \vecX^t\vecC_k$
where $\vecX \in \mathbb{R}^{chan}$ is a feature vector of number
of channels dimension, $y = \{1,-1\}$ stating if the considered
finger is moving or not.   
Once, we have estimated the coefficient vector $\vecC_k$ for each
finger, we selected the $K$ channels that present the largest values
of :
$$
\sum_{k=1}^{6} |\vecC_k|
$$
where the absolute value is considered as element-wise. This 
channel selection allows us to reduce substantially the number
of channels so as to minimize the computational effort
needed for estimating and evaluating the function $f(\cdot)$
and it yields better performance. $K$ has been chosen
so that the cross-correlation on the validation set
is maximal.

\subsubsection{Model estimation}

The model for estimating which finger is moving is
a more sophisticated version of the one
used above for channel selection. At
first, since the finger movements are
mutually-exclusive,  we have considered a winner-takes-all
strategy  :
\begin{equation}
  \label{eq:wta}
  f(\vecX)= \argmax_{k=1,\cdots,6} f_k(\vecX) 
\end{equation}
Here again, $f_k(\vecX)$ is a linear model that is trained
by presenting couples of feature
vector and a state $y=\{1, -1\}$.  
The main differences between the  channel selection procedure and
the one  used for learning $f_k(\cdot)$
are that :  the features here take into account some dynamics
of the ECoG signals and a finer feature selection has been
performed by means of a simultaneous sparse approximation 
method\index{Simultaneous sparse approximation}. \\

Let us consider the training examples
$\{\vecX_t,\vecY_{t}\}_{t=1}^{\ell}$ where $\vecX_t \in \mathbb{R}^d$,
$y_{t,k}=\{1, -1\}$, being the 
$k$-th entry of vector $\vecY_t$, $t$ denoting the time instant and
$k$ denoting all possible states (including no finger moving). 
$y_{t,k}$ tells us whether the finger $k$ is moving 
at time $t$.
Now, let us define the matrix $\matY$, $\matX$
and $\matC$
as :
$$
[\matY]_{t,k}=y_{t,k} \quad [\matX]_{t,j}=x_{t,j}
\quad [\matC]_{j,k}= c_{j,k}
$$
where $x_{t,j}$ and $c_{j,k}$ are 
the $j$-th components of respectively $\vecX_t$ and $\vecC_k$.
The aim of simultaneous sparse approximation is
to learn the coefficient matrix $\matC$ while
yielding the same sparsity profile in the different
finger models. The task boils down to the following
optimization problem:x
\begin{equation}
  \label{eq:ssa}
  \hat \matC = \argmin_{\matC} \|\matY-\matX \matC\|_F^2
+ \lambda_s \sum_{i} \|\matC_{i,\cdot}\|_2 
\end{equation}
where $\lambda_s$ is a trade-off parameter that has
to be appropriately tuned
and $\matC_{i,\cdot}$ being 
the $i$-th row of $\matC$. Note that  our penalty
term is a mixed $\ell_1-\ell_2$ norm similar to
those used for group-lasso. Owing to the 
$\ell_1$ penalty on the $\ell_2$ row-norm, such
a penalty tends to induce row-sparse matrix $\matC$.
Problem (\ref{eq:ssa}) has been solved using the
block-coordinate descent algorithm proposed
by Rakotomamonjy \cite{rakotosparse09}.




\subsection{Learning finger flexion models}
\label{part:fingermodelindent}

\begin{figure}[t]
  \centering
  \includegraphics[width=\linewidth]{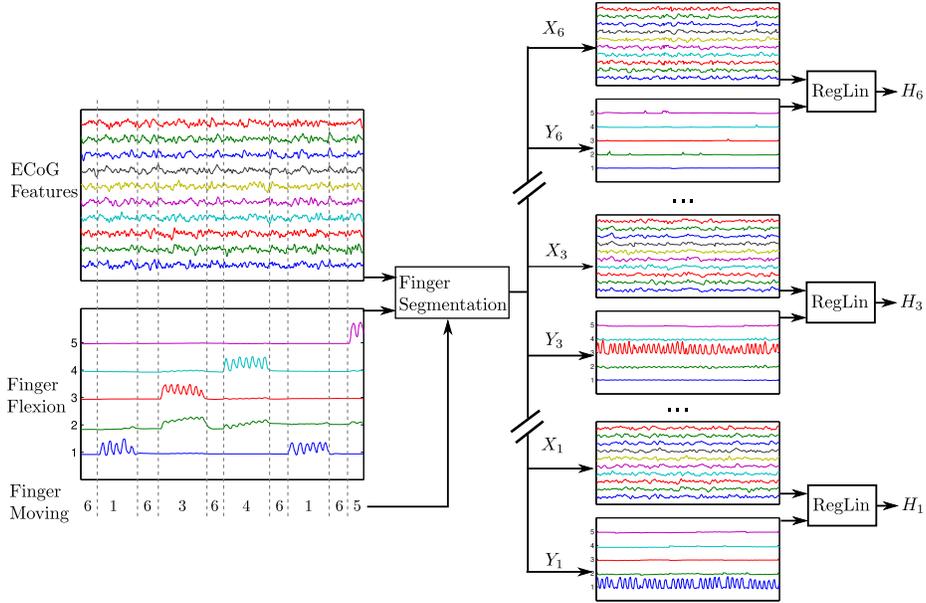}
  \caption{Workflow of the
learning sets extraction ($\matX_k$ and $\matY_k$) and estimation of the linear models $\matH_k$.}
  \label{fig:featlinearmodels}
\end{figure}

Here, we discuss the model relating the ECoG data
and finger movements for every possible values
of $k$. In other words, supposing
that a given finger, say the index, is going to move
(as predicted by our finger moving estimation),
we built an estimation  of all finger movements.
Hence, for each $k$, we are going to learn
a linear model $g_{k,j}(\tilde \vecX)= \tilde \vecX ^T \vecH_j^{(k)} $ with $j=1,\cdots,5$, $\tilde \vecX$ a feature vector
and $\vecH_j^{(k)}$ a weighting vector indexed
by the moving finger $k$ and the finger 
$j$ which flexions are to estimate.
We have chosen a linear model since they have been
shown to provide good performances for decoding
movements from ECoG \cite{Pistohl2008,Schalk2007}. 

At a time $t$, the feature vector
$\vecX_t$ has been obtained by following  the same line as Pistohl et al. \cite{Pistohl2008}. Indeed, we use filtered time-samples as features. 
$\vecX_t$ has been built in the following way.
All channels have been filtered with a Savitsky-Golay (third order, 0.4 s width) low-pass filter. 
Then, $\vecX_t$ is composed of the concatenation
of the time samples at $t$, $t- \tau$ and $t + \tau$
for all smoothed signals at all channels.
Samples at $t- \tau$ and $t + \tau$ have
been used in order to
to take into account some temporal delays between the brain activity
and  finger movements.

Now, let us detail how, for a given moving finger
$k$, the weight matrix $\matH_k = [\vecH_1^{(k)}
\cdots \vecH_5^{(k)}]$ has been learned.
For
a given finger $k$, we have used
as a training set all samples $\vecX_t$ where that finger
is known to be moving. For this purpose, we have manually segmented
the signals and extracted the appropriate signal segments,
needed 
for building the target
matrix $\tilde \matY_k$,
which  
contains all finger positions, and for extracting the feature matrix 
$\tilde \matX_k$. This training samples extraction stage is illustrated
on Figure \ref{fig:featlinearmodels}.
Then for learning the global linear model, we have solved
the following  multi-dimensional ridge regression problem.
\begin{equation}
 \min_{\matH_k}\lVert \tilde \matY_k- \tilde \matX_k \matH_k \rVert^2_F + \lambda_k\lVert \matH_k \rVert^2_F
\label{equ:ridge}
\end{equation} 
with $\lambda_k$ being a regularization parameter that has to
be tuned.

For this problem of finger movements estimation, we also noted
that feature selection helps in improving performance. 
Again, we have used the estimated weighting matrix $\hat \matH$ coefficients
for pruning the model. Indeed, we have
kept in the model the $M$ features which 
correspond to the $M$ largest entries
of  vector $\sum_{i=1}^5|\hat \vecH_i^{(k)}|$. 
For possible $k$ and subjects, $M$ is chosen as to minimize a validation
error. Note that such an approach for pruning model
can be interpreted as a shrinkage of a least-square
parameters.


\subsection{Decoding finger movement}
\label{part:estimatedfingermovement}

When all models have been learned, the decoding
scheme is the one given in Figure \ref{fig:decoder}. 
Given the two feature vectors $\vecX_t$ and $\tilde \vecX_t$
at a time $t$, the finger position estimation is
obtained as:
\begin{equation}\label{fingerpos}
\hat \vecY= \tilde \vecX_t^T \hat \matH_{\hat k} \quad \quad
\text{ with } \hat k= \argmax_k \vecX_t^t\vecC_k
\end{equation}
with  $\hat \vecY$ being a row vector containing
the estimated finger movement, $\hat k$ the finger
that is supposed to move, $\tilde \vecX_t$ the extracted feature at
time $t$
and $\hat \matH_{\hat k}$ the estimated linear model for state $k$.




\section{Results}





In this section we present the performance of our switching
model decoder. At first, we explain how all parameters of
the models have been set. 
Then, we present some results which help us
understanding the contribution of
the different parts of our models,
%
Finally,  we evaluate our approach  and compare ourselves to the BCI
competition results.

\subsection{Parameter selection}
\label{sec:feature-extraction}


The parameters used in the moving finger estimation are selected by a
validation method on the last part of the training set (75,000 for the
training, 25,000 for validation). We suppose that the size of the set
is important enough to avoid over-fitting. Using this method, we
select the number of selected channels, the time-lag $t_s$ used in
feature extraction and the regularization term $\lambda_s$ of
Eq.~\eqref{eq:ssa}.

Similarly, all parameters($\tau$, the number
of selected channels and $\lambda_k$) needed for
estimating $\matH_k$ have been set so that
they optimize the model performance of on
the validation sets. For this model selection
part, the size of training and validation sets
vary according to $k$ and they are
summarized in Table 
 \ref{tab:nbsamplesvalid} for subject 1.


\begin{table}[t]
  \centering
  \begin{tabular}{|c|c|c|}\hline
    Finger & Learning & Validation\\ \hline 
    1 & 8355 & 3848  \\
    2 & 9750 & 2965 \\
    3 & 13794 & 3287 \\
    4 & 6179 & 2915 \\
    5 & 10729 & 5362 \\
    6 & 26074 & 6623 \\ \hline
  \end{tabular}
  \caption{Number of samples used in the validation step for subject 1}
  \label{tab:nbsamplesvalid}
\end{table}




\subsection{Evaluating of the linear models $\matH_k$}

\begin{figure}[t]
  \centering
 \includegraphics[width=\linewidth]{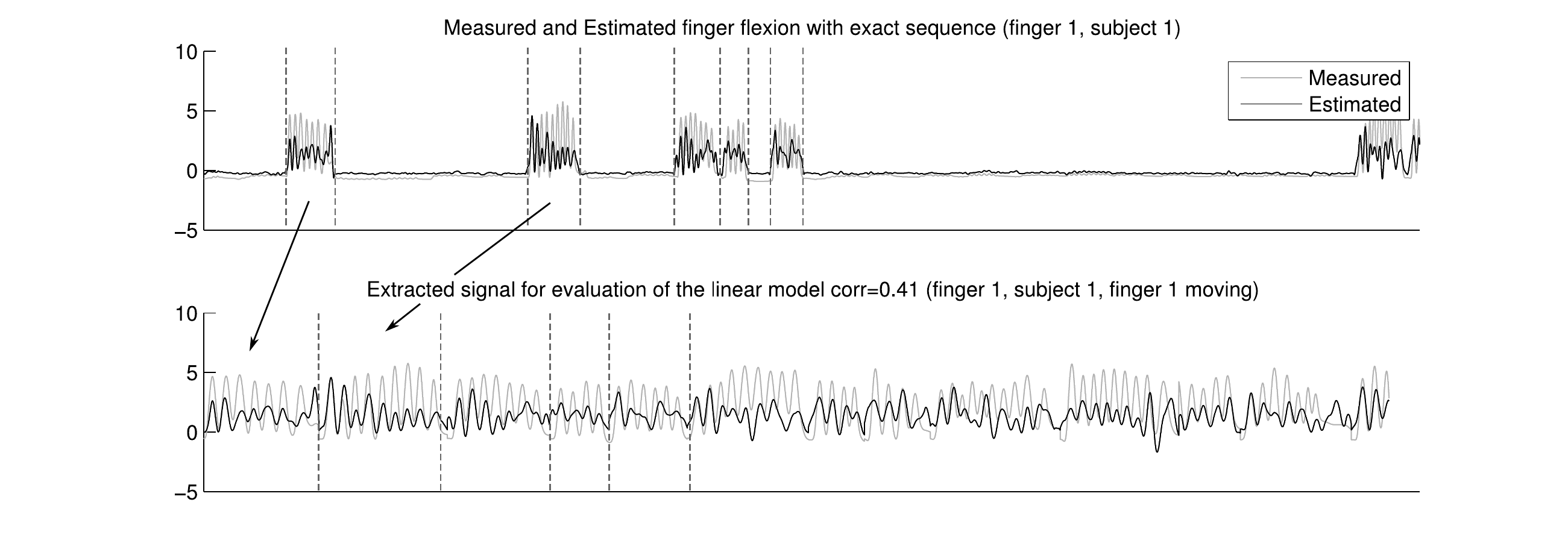}
  \caption{Signal extraction for linear model estimation: 
(upper plot) full
    signal  with segmented signal, corresponding to moving finger, bracketed by the vertical lines
 and (lower plot) the extracted signal 
     corresponding to the concatenation of the samples
    when finger 1 is moving.} 
\label{fig:corrfingerslinear}
\end{figure}

Models $\matH_k$ correspond to the linear regressions between the ECoG
features and the finger flexions when the $k$-th finger is moving.
The signals used for evaluating these models 
are extracted in the same manner as the
learning sets $\vecY_k$ and $ \tilde \matX_k$  
(see Figure \ref{fig:corrfingerslinear})
but by assuming that  the true segmentation
of  finger movements are known.
To evaluate these models, we
measure the cross-correlation between the true
$\vecY_k$
and estimated
finger flexion $\tilde \matX_k \hat \vecH_k^k$
only when the finger 
$k$ is moving.
The correlations can be seen
on Table \ref{tab:corrfingerslinear}.
\begin{table}[t]
 
  \begin{center}
    \begin{tabular}{|l||c|c|c||c|} \hline Finger & Sub. 1 & Sub. 2 &
      Sub. 3 & Average\\ \hline
      1&0.4191&0.5554&0.7128&0.5625\\
      2&0.4321&0.4644&0.6541&0.5169\\
      3&0.6162&0.3723&0.2492&0.4126\\
      4&0.4091&0.5668&0.0781&0.3513\\
      5&0.4215&0.5165&0.5116&0.4832\\ \hline
      Avg.&0.4596&0.4951&0.4411&0.4653\\\hline
    \end{tabular}
  \end{center}

  \caption{Correlation coefficient 
obtained by the linear models 
$\vecH_k^{(k)}$.  }
  \label{tab:corrfingerslinear}

\end{table}

We observe that by using a linear regression between the ECoG signals
and the finger flexions, we achieve a correlation of
0.46 (averaged across fingers and subjects).
This results correspond to those obtained for the arm trajectory
prediction (Schalk \cite{Schalk2007} obtained 0.5 and Pistohl
\cite{Pistohl2008} obtained 0.43). 

\subsection{Evaluating  the switching decoder method}
\label{sec:eval-switch-models}

In order to evaluate the efficiency of the switching model decoder
and  each block of the decoder contribution.
We report three different results: first,
for a given finger,  we compute the estimated 
finger flexion  using a linear model
learned on all samples (including those where the
considered finger is not moving), then we decode finger flexions
with our switching
decoder while assuming that the exact sequence of hidden states 
is known\footnote{This is possible since
the finger movements on the test set are now available} and finally we use
our switching decoder with the estimated hidden states.

\begin{figure}[t]
 \centering
\subfloat[Subject 1, Finger 1]{\label{fig:perfect}
 \includegraphics[width=0.5\linewidth]{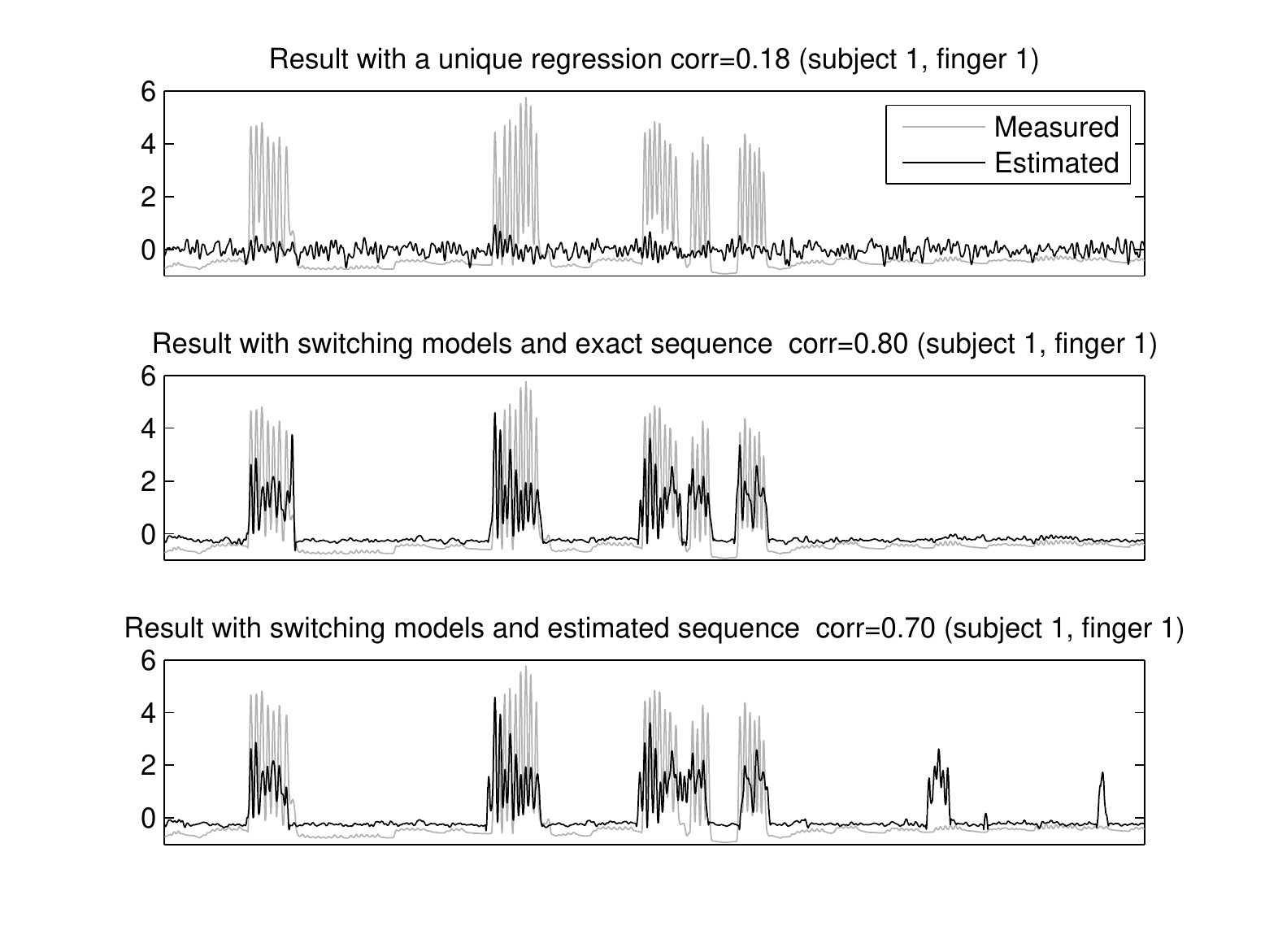}}
\subfloat[Subject 1, Finger 2]{\label{fig:est}
 \includegraphics[width=0.5\linewidth]{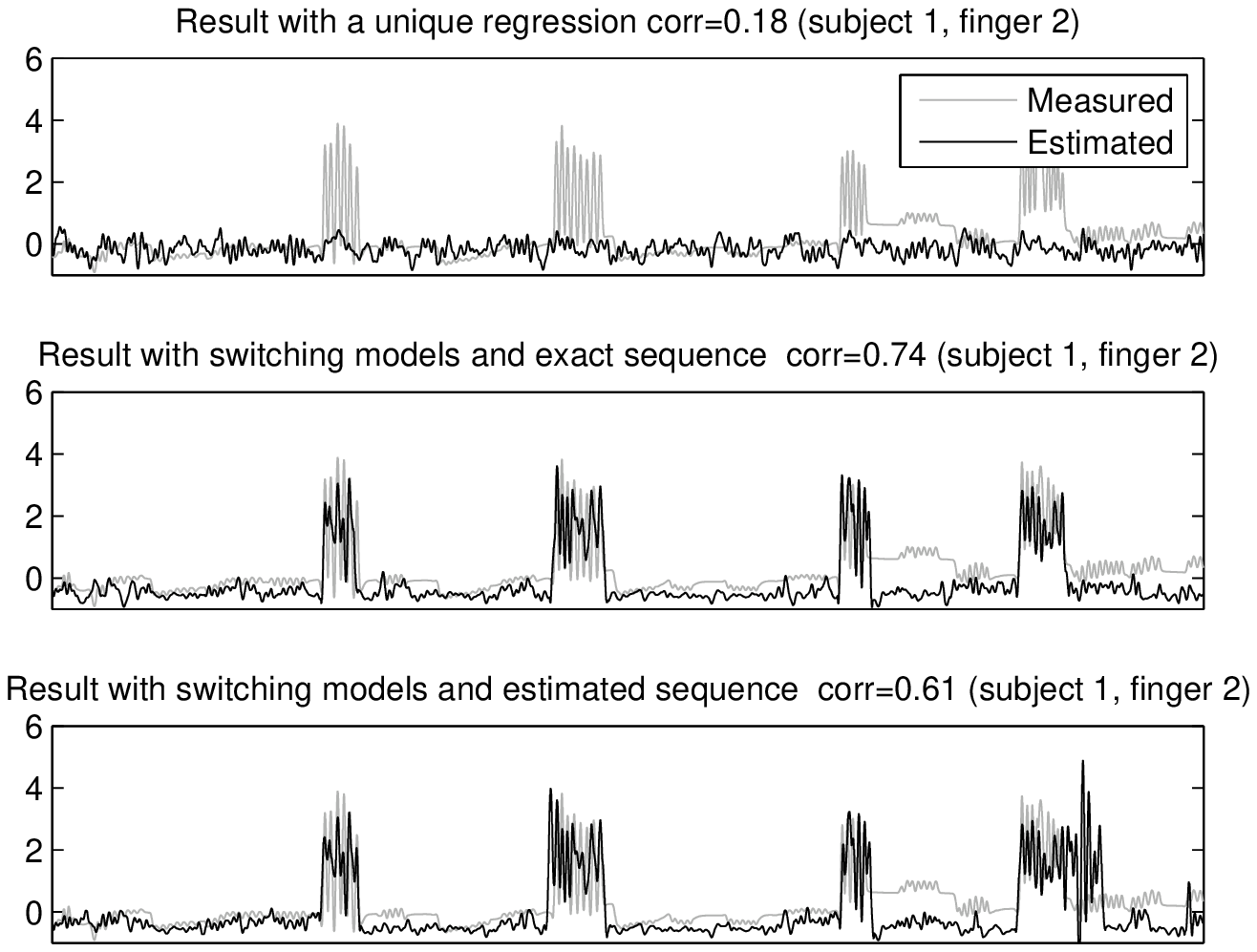}}
 \caption{True and estimated finger flexion for  (upper plots) a global  linear
   regression,   (middle plots) switching decoder with true moving
finger segmentation
  and (lower plots) with the switching decoder  with an estimated
   moving
finger segmentation. } \label{fig:visures}
\end{figure}

For a sake of baseline comparison with our 
switching models decoder, we have  estimated
the finger flexions by means of a single linear model
which has been trained using all the time samples.
The obtained correlation are given in
Table \ref{tab:linear} and the regression result can be seen on the upper
plots of Figure \ref{fig:visures}. We can see that the correlation obtained are
rather low due the fact that without switching models the amplitude of
the flexion signals remains small.

The switching model decoder is a two-part process as it requires to have the
linear models $\matH_k$ and the sequence of hidden states. First we
apply the decoder using the true sequence obtained thanks to the
actual finger flexion.
Suppose that we have the exact sequence $\textbf{k}$
and we apply the switching decoder with this sequence. We know that
these results may never be attained as it supposes the sequence labeling
method to be perfect but it gives a interesting idea of the maximal 
performance
that our method can provide for given linear models $\matH_k$. Results can
be seen in the middle plots of Figure \ref{fig:visures} and correlations
are in Table \ref{tab:exact}.
We obtain a high accuracy accross all subjects with an average
correlation of 0.61 when using an exact sequence. 
This proves that the switching model can be efficiently used  for
decoding ECoG signals.
Note that by using switching linear models, we
include a switching mean that induce a high accuracy of correlation.


\begin{table}[ht]
 \begin{center}
\begin{footnotesize}
\end{footnotesize}
\subfloat[Linear regression]{\label{tab:linear}
  \begin{tabular}{|l||c|c|c||c|}\hline Finger & Sub. 1 & Sub. 2 &
      Sub. 3 & Average\\ \hline
    1&0.1821&0.2604&0.3994&0.2807\\
2&0.1844&0.2562&0.4247&0.2884\\
3&0.1828&0.2190&0.4607&0.2875\\
4&0.2710&0.4225&0.5479&0.4138\\
5&0.1505&0.2364&0.3765&0.2545\\ \hline
Avg.&0.1942&0.2789&0.4419&0.3050\\ \hline
  \end{tabular}
}
\subfloat[Switching models (exact sequence)]{\label{tab:exact}
  \begin{tabular}{|l||c|c|c||c|}\hline Finger & Sub. 1 & Sub. 2 &
      Sub. 3 & Average\\ \hline

1&0.8049&0.5021&0.8030&0.7033\\
2&0.7387&0.4638&0.7655&0.6560\\
3&0.7281&0.4811&0.7039&0.6377\\
4&0.7312&0.5366&0.6241&0.6307\\
5&0.2296&0.4631&0.6126&0.4351\\ \hline
Avg.&0.6465&0.4893&0.7018&0.6126\\ \hline
  \end{tabular}
}\\
\subfloat[Switching models (est. sequence)]{\label{tab:estim}
  \begin{tabular}{|l||c|c|c||c|}\hline Finger & Sub. 1 & Sub. 2 &
      Sub. 3 & Average\\ \hline
1&0.7016&0.3533&0.6457&0.5669\\
2&0.6129&0.3045&0.5097&0.4757\\
3&0.2774&0.0043&0.4025&0.2280\\
4&0.4576&0.2782&0.5920&0.4426\\
5&0.3597&0.2507&0.6553&0.4219\\ \hline
Avg.&0.4818&0.2382&0.5611&0.4270\\ \hline
  \end{tabular}
}
\end{center}
\caption{Correlation between measured and estimated movement for a global linear
   regression (a), switching decoder with exact sequence
   (b) and switching decoder with an estimated sequence (c)}
\label{tab:corr}
\end{table} 


Finally, we use our global  method for obtaining the finger movement
estimation. In other words, we used the switching models $\matH_k$ to decode
the signals with Equation (\ref{fingerpos})
 and the estimated sequence $\hat{\mathbf{k}}$.
The finger movement estimation  can be
seen on the lower plot of Figure \ref{fig:est} and the correlation
measures are in Table \ref{tab:estim}. As expected, the accuracy is
lower than the one obtained with the
 true segmentation. However, we obtained an average
correlation of $0.42$ which is far better than when using a 
global regression approach. 
These predictions of the finger flexions were
presented in the BCI Competition and achieved the second place. 
 Note that the last 3 fingers have the lowest
correlation. Those one are highly physically correlated and they are much
more difficult to discriminate than the two first ones. The first
finger is by far the best estimated one as we obtained a correlation
averaged accross subject of 0.56 .

\subsection{Discussion and future works}

The results presented in the previous section corresponds to the
method used for the BCI Competition.

We first note that the
best performance obtained by  Liang et al. \cite{bougrainesann09}
 gives  a correlation of about $0.46$. Their
method considers an amplitude modulation
along time to cope with the abrupt change in the finger flexions
amplitude along time. Such an approach is somewhat similar to
ours since they try to distinguish situations where fingers
are moving or not.

Then, we believe that our approach can be improved in
several ways. 

Indeed, we choose to use linear models depending on the internal
states, but \cite{Pistohl2008} proposed to use a kalman filter for the
decoding of movement. This approach may be extended to using switching
kalman filters in the switching model decoder.

Furthermore, our approach for estimating the sequence of hidden states can be
highly improved. Liang~\cite{bougrainesann09} proposed to use Power
Spectral Densities of the ECoG channel as features and we believe that
these features may be added and used in the sequence
labeling. 
In our method the features are low-pass filtered in order
to increase recognition performance, but other sequence labeling
methods like HMM \cite{darmajian06} have been used in BCI. Other
sequence labeling methods like Conditional Random Fields
\cite{Lafferty01Conditional} known to outperform HMM in some case or
Sequence SVM \cite{bordes-usunier-bottou-2008} may be used to get a
better sequence of hidden states.

Another interesting approach that may be investigated
is the mixture of sources approach. Indeed, one may 
considered that each moving finger is associated
to a source of ECoG signals. Then, the problem of
identifying which finger is moving may boil down to
a source separation problem.



\section{Conclusions }

In this paper, we present a method for the decoding  finger flexions
from ECoG signals. 
The decoder is based on switching linear models.
Our approach has been tested on a the BCI Competition IV
Dataset 4 and achieved the second place in the 
competition.
Results show that the switching model approach produce better result than using
a unique model. Furthermore an
accurate finger flexion estimation  may be 
achieved  when using an exact sequence of
hidden states showing the interest of the switching models.

In future works, we plan to improve the result of the switching models
decoder by two different approaches. On the one hand, we can use more
general models than linear ones for the movement prediction (switching
kalman filters,
non-linear regression). On the
other hand we can improve the sequence labeling along time with new
approach and by using new features extracted from the ECoG signals.


\bibliographystyle{splncs}
\bibliography{biblioRF,biblioAR}

\end{document}